\documentclass[manuscript,screen]{acmart}
\AtBeginDocument{%
  \providecommand\BibTeX{{%
    \normalfont B\kern-0.5em{\scshape i\kern-0.25em b}\kern-0.8em\TeX}}}

\setcopyright{acmcopyright}
\copyrightyear{2021}
\acmYear{2021}
\acmDOI{XXXXXXX.XXXXXXX}

\begin{document}

\title{\textbf{BankNote-Net}: Open dataset for assistive universal currency recognition}

\author{Felipe Oviedo}
%% \authornote{Both authors contributed equally to this research.}
\email{felipe.oviedo@microsoft.com}
%%\orcid{1234-5678-9012}
\affiliation{%
  \institution{Microsoft AI for Good Research Lab}
  \streetaddress{1 Microsoft Way}
  \city{Redmond}
  \state{WA}
  \country{USA}
  \postcode{98052}
}

\author{Srinivas Vinnakota}
\affiliation{%
  \institution{Microsoft}
  \streetaddress{1 Microsoft Way}
  \city{Redmond}
  \state{WA}
  \country{USA}
  \postcode{98052}
}

\author{Eugene Seleznev}
\affiliation{%
  \institution{Microsoft Seeing AI}
  \streetaddress{1 Microsoft Way}
  \city{Redmond}
  \state{WA}
  \country{USA}
  \postcode{98052}
}

\author{Hemant Malhotra}
\affiliation{%
  \institution{Microsoft}
  \streetaddress{1 Microsoft Way}
  \city{Redmond}
  \state{WA}
  \country{USA}
  \postcode{98052}
}

\author{Saqib Shaikh}
\affiliation{%
  \institution{Microsoft Seeing AI}
  \streetaddress{1 Microsoft Way}
  \city{Redmond}
  \state{WA}
  \country{USA}
  \postcode{98052}
}

\author{Juan Lavista Ferres}
\affiliation{%
  \institution{Microsoft AI for Good Research Lab}
  \streetaddress{1 Microsoft Way}
  \city{Redmond}
  \state{WA}
  \country{USA}
  \postcode{98052}
}

\begin{abstract}

Millions of people around the world have low or no vision. Assistive software applications have been developed for a variety of day-to-day tasks, including optical character recognition, scene identification, person recognition, and currency recognition. This last task, the recognition of banknotes from different denominations, has been addressed by the use of computer vision models for image recognition. However, the datasets and models available for this task are limited, both in terms of dataset size and in variety of currencies covered. In this work, we collect a total of 24,826 images of banknotes in variety of assistive settings, spanning 17 currencies and 112 denominations. Using supervised contrastive learning, we develop a machine learning model for universal currency recognition. This model learns compliant embeddings of banknote images in a variety of contexts, which can be shared publicly (as a highly-compressed vector representation), and can be used to train and test specialized downstream models for any currency, including those not covered by our dataset or for which only a few real images per denomination are available (few-shot learning). We deploy a variation of this model for public use in the last version of the \href{https://www.microsoft.com/en-us/ai/seeing-ai}{Seeing AI} app developed by Microsoft. We share our encoder model and the embeddings as an open dataset in our \href{https://github.com/microsoft/banknote-net}{BankNote-Net repository}.

\end{abstract}




\maketitle

\section{Introduction}

According to recent data \cite{bourne2021trends}, 43 million individuals worldwide are estimated to suffer blindness and over 200 million people suffer moderate or severe vision impairment. 55\% of people in this group are women and 89\% live in low and middle-income countries. Due to population growth and higher incidence of certain pathologies \cite{bourne2021trends}, the number of people affected by blindness and moderate to severe vision loss is expected to increase to over 61 million and 235 million people by 2050, respectively. 

Given the magnitude of the problem, the software community has developed numerous mobile applications and wearable technologies to aid visually impaired individuals in a multitude of tasks \cite{manduchi2018assistive, tapu2020wearable, hakobyan2013mobile}, including navigation, public transport use, scene understanding, person recognition, currency note recognition, etc. Seeing AI \cite{SeeingAI}, a major application in the space developed by Microsoft, is a pioneer of applying AI to assist individuals with low and no vision and currently has 100 thousand average monthly users in 80 countries, supporting vision-to-speech in 19 languages. One common task of assistive applications consists in the recognition (\textit{i.e.} multi-class classification) of banknotes of different denominations for a specific currency. Machine learning (ML) is often applied for this task. Several ML models have been develop for particular currencies \cite{raval2019comparative}, including Indian Rupee \cite{veeramsetty2020coinnet, upadhyay2020indian, jain2021automated, rahman2014lda}, US dollar \cite{hasanuzzaman2012robust, gai2013employing, paisios2011recognizing}, Canadian dollar \cite{paisios2011recognizing}, Euro \cite{raval2018horbovf}, Bangladesh Taka \cite{islam2021real}, Thailand Baht \cite{takeda2003thai}, Egyptian Pound \cite{semary2015currency}, Pakistan Rupee \cite{imad2020pakistani}, among others \cite{raval2019comparative}. The techniques employed in previous works include both classical computer vision and machine learning approaches, such as LDA \cite{rahman2014lda}, feature transformations and template-matching \cite{raval2018horbovf, gai2013employing, semary2015currency}, support vector machines \cite{upadhyay2020indian}, and modern deep learning approaches \cite{islam2021real, veeramsetty2020coinnet, takeda2003thai, imad2020pakistani}.

Either for training or for testing purposes, these models rely on a relatively large dataset of labelled banknote images spanning all the denominations of certain currency. If the model is going to be deployed in a mobile assistive application, as is common for the banknote recognition task, production quality models require well-curated datasets in a variety of real assistive scenarios, including images with poor focus, various backgrounds, occlusions, blur, varying orientation and perspective, etc. The lack of comprehensive public datasets has made limited the development of these models to a few particular countries and currencies. 

\begin{figure}[h]
  \centering
  \includegraphics[width=400pt]{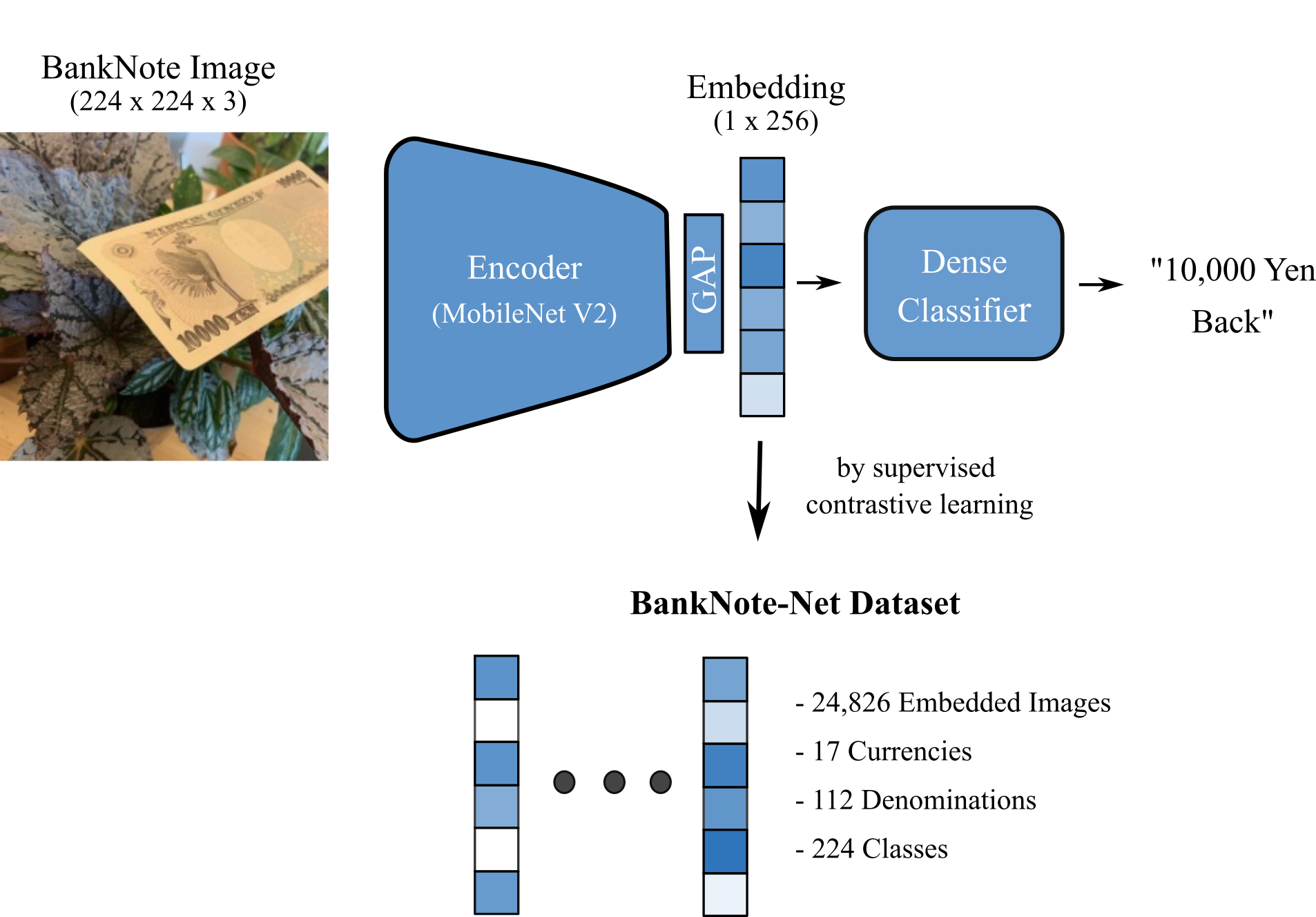}
  \caption{\textbf{Our approach}: we use a convolutional neural network (MobileNet V2) as an encoder for our assistive banknote images. Using supervised contrastive learning, the models learns an highly-compressed and descriptive embedding for each image. We use these embeddings for downstream tasks, such as recognition of Yen banknotes, as shown in the figure. Our final dataset is composed of the learned embeddings for 17 currencies and 224 classes. "Dense Classifier" corresponds to a fully-connected neural network, and "GAP" corresponds to a Global Average Pooling layer.} \label{fig:approach}
\end{figure}

In this work, our goal is to facilitate the training of machine learning models for the recognition of any banknote in assistive settings. For this purpose, we collect, curate and label BankNote-Net: the largest open dataset of banknote images for accessibility to date, composed of a total of 24,826 images of front and back banknote faces across 17 currencies and 112 denominations. To approximate its use by an individual with low or no vision, the images are captured in a variety of orientations, lightning, background, occlusion and image quality conditions. Using supervised contrastive learning \cite{khosla2020supervised}, we use this dataset to train a universal deep learning model for multi-class classification spanning the 17 currencies and 224 classes (2 faces per each one of the 112 denominations). As the reproduction of high-resolution currency images is regulated in some jurisdictions, such as as England \cite{Usingima22} or Japan \cite{Lawsconc97}, we use this model to learn highly compressed and compliant embeddings for the images. Using leave-one-group-out cross validation, we demonstrate these embeddings are useful for pre-training models for currencies beyond those covered by our dataset for both the abundant data setting and the few-shot learning case. Figure \ref{fig:approach} summarizes our approach. These embeddings, along with the pre-trained encoding model, are shared as a public dataset in our \href{https://github.com/microsoft/banknote-net}{BankNote-Net repository}. In addition, we use our dataset for successful validation and deployment of a variation of our model in production, as part of the most recent version of the Seeing AI app \cite{SeeingAI}.

\section{Data collection and curation}

Deploying reliable ML models for banknote recognition requires images of banknotes in real assistive scenarios, \textit{i.e.} varying background, illumination, blur, orientation, occlusion and focus conditions. Although, high quality scans of multiple currencies are available in banknote collectors' websites such as \cite{BankNotews, IBNS}, abundant data of real usage scenarios for assistive applications is very limited. In consequence, we design our data collection process with this accessibility application in mind. 

The final data collection process consisted of the following steps:

\begin{enumerate}
    \item \textbf{Currency selection}: we selected the particular currencies to collect according to population and geographic diversity, and according to areas with high number of users of the Microsoft Seeing AI app. 
    \item \textbf{Recruitment}: for each specific currency, we on-boarded consenting volunteers or Amazon Mechanical Turk \cite{MTurk} workers to perform data collection.
    \item \textbf{Training}: each worker or volunteer was provided with a brochure of requirements for the desired pictures along with example pictures. This brochure is included in Appendix \ref{appendix-a}.
    \item \textbf{Data collection}: each worker or volunteer captured and labelled up around 100 images for each orientation (front or back) of each denomination of a specific currency. Each picture was captured with a square aspect ratio to facilitate preprocessing. The images of each currency were captured exclusively by 1 or 2 volunteers or workers located in the country where a specific currency is used; this decision was made to improve data completeness across all denominations of a given currency.
    \item \textbf{Data validation and preparation}: the collected data was validated manually by the authors to check for inconsistencies in labelling or limited image diversity. Then, the file formats were standardized and all images were resized to 224 x 224 pixels.
\end{enumerate}

\begin{figure}[h!]
  \centering
  \includegraphics[width=400pt]{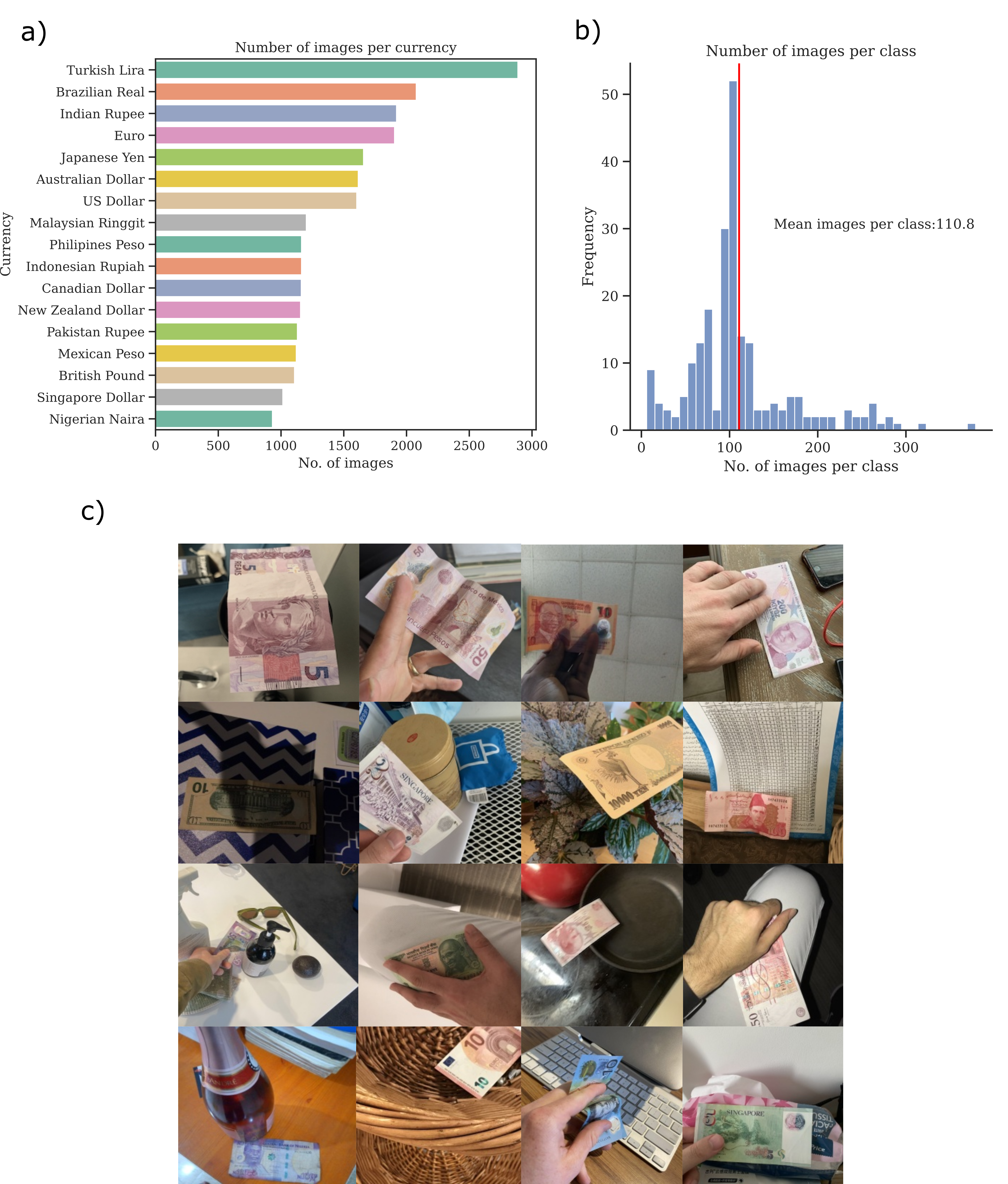}
  \caption{\textbf{Overview of BankNote-Net}: a) Total number of images per currency. b) Distribution of number of images per class (combination of denomination and front/back side, \textit{e.g.} "5 USD - Front Side" is a class). c) Example of images in the dataset, collected in diverse accessibility scenarios.} \label{fig:allimages}
\end{figure}

In total, 24,826 images were collected from 17 currencies. Figure \ref{fig:allimages}a presents the distribution of collected images across currencies, the distribution of images per class, and multiple examples of the collected images. The images span 112 denominations and 224 classes (each denomination maps to two classes, corresponding to the front and back faces of the banknote). The mean number of images per class is 110, as shown in Figure \ref{fig:allimages}. For banknotes that have more than a one version in circulation, we defined each banknote version as a different class. Appendix \ref{appendix-b} has a list of all denominations and classes in our dataset.

\section{Data Representation and Evaluation}

The BankNote-Net dataset can be used for various computer vision tasks, including fine-tuning of recognition models for those currencies included in the dataset, robust validation of recognition models, and transfer learning and few-shot learning to novel currencies. For this purpose, we encode all BankNote-Net images as embeddings and make them publicly available. In the following subsections, we present our encoder model and demonstrate the utility of our embeddings.

\subsection{Learning banknote representations}

We are interested in learning a compact and compliant representation of banknote images that can be used for downstream image recognition tasks. For this purpose, we train a neural network image encoder using a supervised contrastive learning loss \cite{khosla2020supervised}. The supervised contrastive learning loss combines a contrastive loss with a cross entropy loss to learn improved discriminative representations across all classes. Since most assistive applications are deployed in mobile devices, we use the MobileNet V2 architecture \cite{sandler2018mobilenetv2} ($\alpha=1.3$, global average pooling for feature extraction, Image-Net weights, temperature for supervised contrastive loss T=$0.05$, ADAM optimizer with $\textit{lr}=5e^{-5}$ and $\textit{epochs}=150$) as the encoder backbone, as shown in Figure \ref{fig:approach}. The encoder outputs a 256-dimensional vector embedding $v \in \mathbb{R}^{1\times 256}$. During training with the supervised contrastive loss, we find beneficial to freeze the bottom layers in the backbone architecture up to layer $n=96$. $n$, along all other discussed hyper-parameters, are chosen by hyperparameter tuning of the mean AUC across all classes on a 10\% random validation set,  using the Optuna package \cite{akiba2019optuna}. During training, we use random rotation, cropping and channel shift to augment the data.

For each image in our dataset, this model returns a highly-informative and compact embedding, which can be used for downstream tasks, such as multi-class classification, transfer learning or few-shot learning. Our machine learning model is summarized in Figure \ref{fig:approach}.

\subsection{Leave-currency-out cross validation}

We perform a leave-one-group-out cross validation procedure to confirm the predictive power of our embeddings. For this, we hold out one of the 17 currencies in the dataset at a time, and train our encoder with the remaining currencies. For each held-out currency, we split data according to two kinds of splits that we called: \textit{full split} and \textit{few-shot split}. The \textit{full split} corresponds to a 80\%-20\% random train-test split, representing a setting with ~500-1500+ images per denomination. The \textit{few-shot split} corresponds to a 10\%-90\% train-test split, representing a few-shot learning setting with ~5-20 training images per class.

For each one these held-out data splits, we train an encoder model and use it as a feature extractor for the testing currency. On top of this model, we train a classification head consisting of a global average pooling layer and 2 fully connected layers ($\textit{dropout rate}=0.5$, Adam optimizer, $\textit{lr}=1e-03$,  $\textit{epochs}=50$). For each test currency, we average the results over 5 random seeds. It is important to emphasize that the test currency is never used during the training and hyperparameter tuning of the encoder. We evaluate three different encoder initialization scenarios:

\begin{itemize}
    \item \textbf{BankNote-Net}: Pre-training encoder with Banknote-Net embeddings of all training currencies, leaving out testing currency. Frozen encoder, training and testing of top layers on test currency.
    \item \textbf{ImageNet}: Pre-training encoder with ImageNet. Frozen encoder, training and testing of top layers on test currency.
    \item \textbf{Random}: Training of encoder and top layers from scratch, all layers trainable.
\end{itemize}

\begin{figure}[!h]
    \centering
    \includegraphics[width=0.6\textwidth]{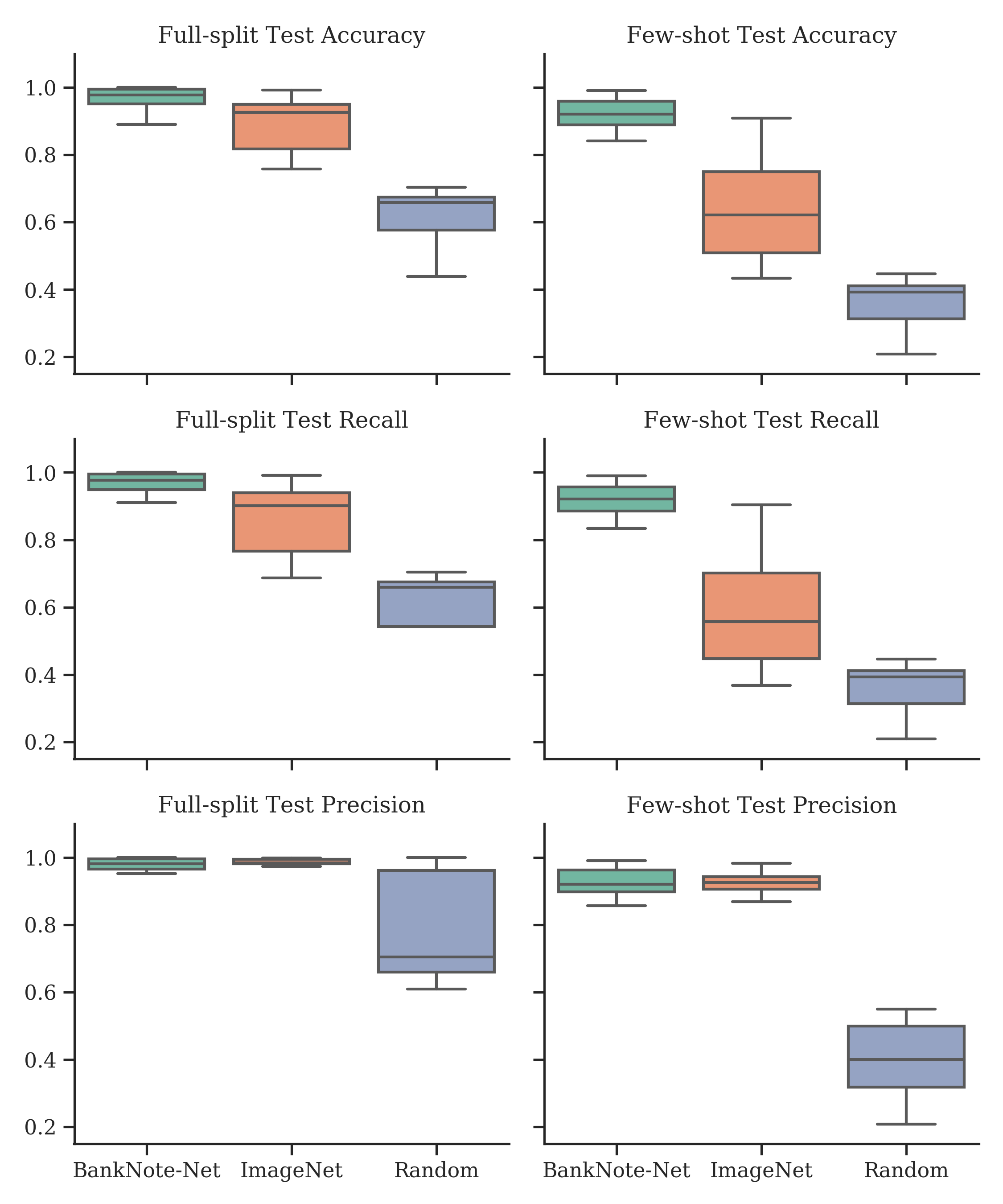}
    \caption{Distribution of accuracy, recall and precision for all test currencies after leave-currency-out cross validation. We use a particular initialization (BankNote-Net, ImageNet, Random) and a specific train-test split (Full-split or Few-Shot split).}
    \label{fig:accuracy}
\end{figure}

Figure \ref{fig:accuracy} compares the accuracy, recall and precision for the three initialization scenarios and the two kinds of data splits. We observe that the Banknote-Net embeddings lead to superior and less variable performance in both the full  and few-shot splits. In the few-shot split, we observe a notable boost in performance, confirming the generalization power of our approach to unseen currencies, even when data is scarce. We note that the accuracy and recall using BankNote-Net weights outperform those models that use other initialization, including ImageNet. The precision for the BankNote-Net initialization is comparable to ImageNet, and is significantly superior to training the model from scratch. The performance on unseen currency data is comparable to models trained exclusively on a single currency, such as \cite{jain2021automated, imad2020pakistani, islam2021real, paisios2011recognizing}.

\subsection{Public embeddings}

As discussed, the reproduction of high-resolution banknote images is regulated in some jurisdictions, such as as England \cite{Usingima22} or Japan \cite{Lawsconc97}. Thus, we make our dataset public in the form of embeddings, which are distilled and compliant representations of the banknote images. Although our original images are only 224x224 pixels in size and are not high resolution scans of banknotes, we provide the following analysis for the benefit of the reader.

Our embeddings can be seen as a \textit{lossy} hash of the collected images. To empirically demonstrate this, we use Shannon coding theorem \cite{cheng2020does, shannon2001mathematical}. According to the theorem, the maximum lossless compressed image size $C$ using pixel-wise 8-bit compression is:

\begin{equation}
    C \leq S \: \frac{H}{8}
\label{shannon}
\end{equation}

Here, S is the original image size (in bytes) and H is the Shannon entropy of an image. 

Modern lossless image compression algorithms, such as LZMA and DEFLATE, exploit the local structure in a image to surpass the limit defined by Equation \ref{shannon} \cite{cheng2020does}. If the image compression algorithm is allowed to be lossy, as in JPEG, it can be further compressed; however, the image will not be able to be decompressed without resolution losses. Our embedding technique compresses the image significantly way beyond commonly used lossy compression algorithms, which in turn makes high resolution reconstruction of an image from the embedding infeasible. 

We illustrate this fact in Figure \ref{embedding}. In the top figure we present the t-SNE representation for the embeddings of 5 major currencies in the dataset. The points naturally cluster around currencies, which confirms the discriminative power of our embeddings. In the bottom figure, we present the original file size (in kilobytes (KB), before downsizing to 224x224 pixels) of each image in the dataset according to its entropy. The Shannon coding theorem sets a theoretical limit for the compressed data points as described in Equation \ref{shannon}. As discussed, lossless JPEG surpasses this limit, specially for cases with low image entropy. The 50\% lossy JPEG image improves on the lossless JPEG compression, with most images in the dataset located below the Shannon coding bounds. Our BankNote-Net embeddings surpass the best compression ratio by 1-2 orders of magnitude. This significant compression level, combined with the initial image resizing step, makes lossless and high resolution reconstructions of the images in the dataset infeasible.

\begin{figure}[!h]
    \centering
    \label{fig:embedding}
    \includegraphics[width=0.7\textwidth]{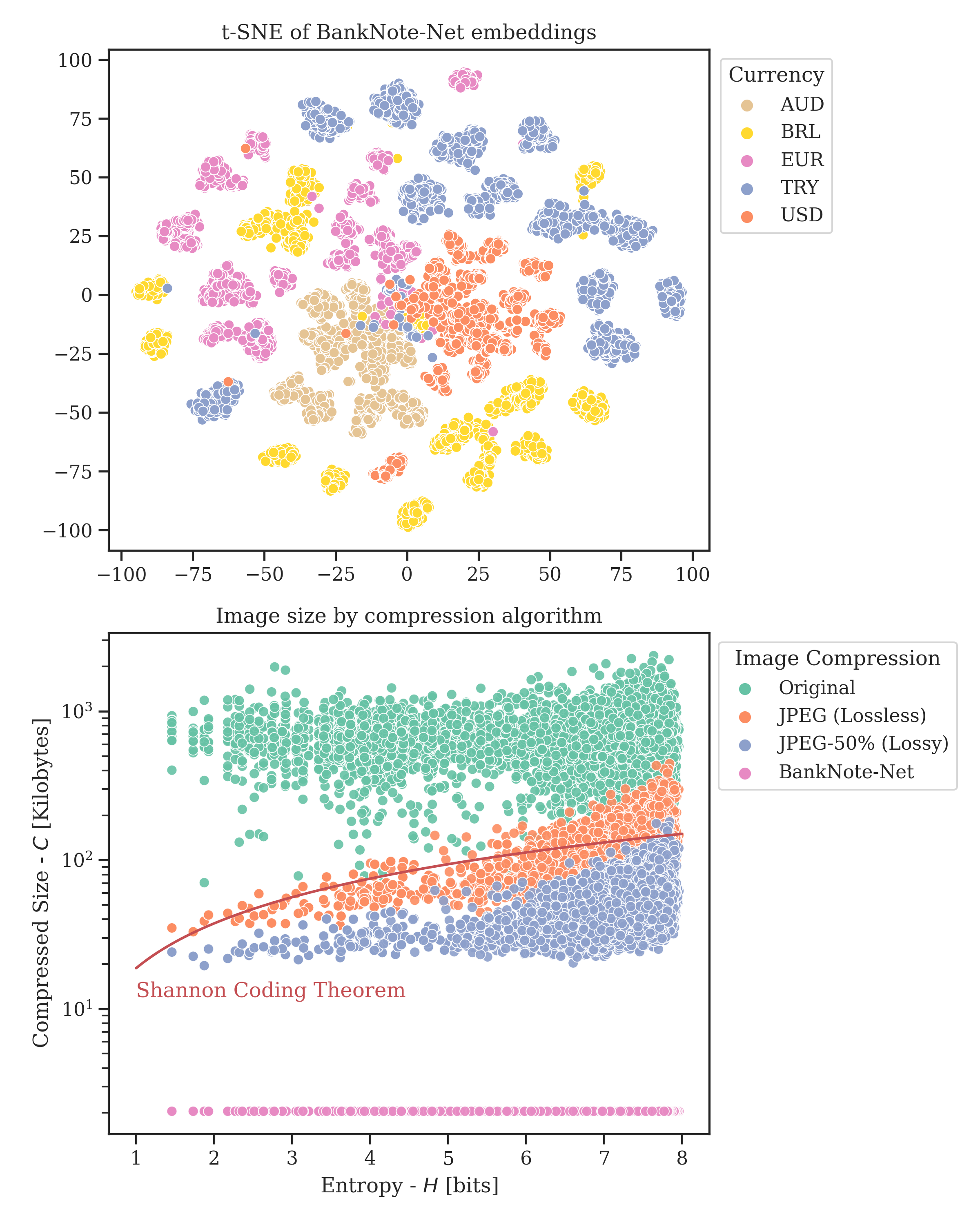}
    \caption{Top: t-SNE representation of the embeddings of 5 major currencies in the BankNote-Net dataset. The embeddings naturally cluster individual currencies together, aiding predictive performance of models learned on the BankNote-Net data. Other currencies in the dataset are omitted for plot readability. Bottom: Compressed size (in Kilobytes) for each image in the dataset as a function of entropy. Our encoded (BankNote-Net) representation is several orders of magnitude smaller than lossy and lossless compression algorithms, making high-resolution reconstruction infeasible.}
    \label{fig:embedding}
\end{figure}

\section{Open Dataset and Production Deployment}
\label{deployment}

Originally, the Seeing AI app supported recognition of 7 currencies that were deployed and serviced as independent models for each currency. With Banknote-Net, we substantially expanded the currencies covered by the Seeing AI app, by using the dataset to train and validate a variation of the universal model for all 17 supported currencies. In this case, we perform further hyperparameter tuning to maximize the model precision for predictions with confidence of 0.99 or above. We choose this approach to minimize false positives during assistive use. This single universal model is deployed in the last version of the \href{https://www.microsoft.com/en-us/ai/seeing-ai}{Seeing AI} mobile app.

The embeddings, labels and the pre-trained MobileNet V2 encoder model developed during work are available in the \href{https://github.com/microsoft/banknote-net}{BankNote-Net repository}.

\section{Discussion and Conclusions}

In this work, we collect and make public a large dataset of banknote images for assistive recognition. To share the dataset, we use supervised contrastive learning to get learn compliant and highly-compressed embeddings for each of the 24,826 images and the 17 currencies in the dataset. We demonstrate the power of these embeddings to improve the predictive performance of currency recognition models for assistive applications, including the few-shot scenario. Additionally, we demonstrate that the embeddings cannot be feasibly \textit{decompressed} to high-resolution images.

Our dataset has several limitations to consider. In particular, most of the images of a given currency were collected by 1-2 volunteers or workers. Although the workers were clearly aiming to optimize diversity of settings, backgrounds and orientations for the images, the final diversity of the dataset can be limited by bias during data collection, inherent specifications of the mobile camera employed, limited background for image capture, etc. This constitutes an important trade-off we made for the data collection process, and we chose to perform it in this way to maximize completeness across the denominations within a given currency. Regardless, we show that the learned embeddings have substantial generalization power, including to currencies not covered by the BankNote-Net dataset. This makes the dataset very useful for future applications, for instance to extend the model to non-supported currencies or to include new versions of banknotes that were not considered in the dataset. In the future, we may consider expanding the dataset to achieve a broader diversity of data collectors.

Another important consideration is that the deployment of the model in production may require the optimization of particular metric of interest beyond precision and recall, \textit{e.g.} the high confidence precision mentioned in Section \ref{deployment}. In this case, the dataset user may benefit from re-training or fine-tuning the encoder on a representative validation dataset to achieve good performance on the custom metric. 

\begin{acks}
The authors thank Ria Sankar and Heila Precel for helpful discussion. We acknowledge the contributions of all data collectors across the world.
\end{acks}

\section{Data and Code Availability}
The BankNote-Net dataset, along with an encoder, labels and usage examples is available as an open dataset under the MIT and CDLA-Permissive-2.0 licenses in the \href{https://github.com/microsoft/banknote-net}{BankNote-Net repository}.

\bibliographystyle{ACM-Reference-Format}
\bibliography{references}

\appendix
\section{Appendix A} \label{appendix-a}

The data collection instructions and image examples shared with data collectors are included below.
\begin{figure}[H]
\centering
\includegraphics[width=0.7\textwidth]{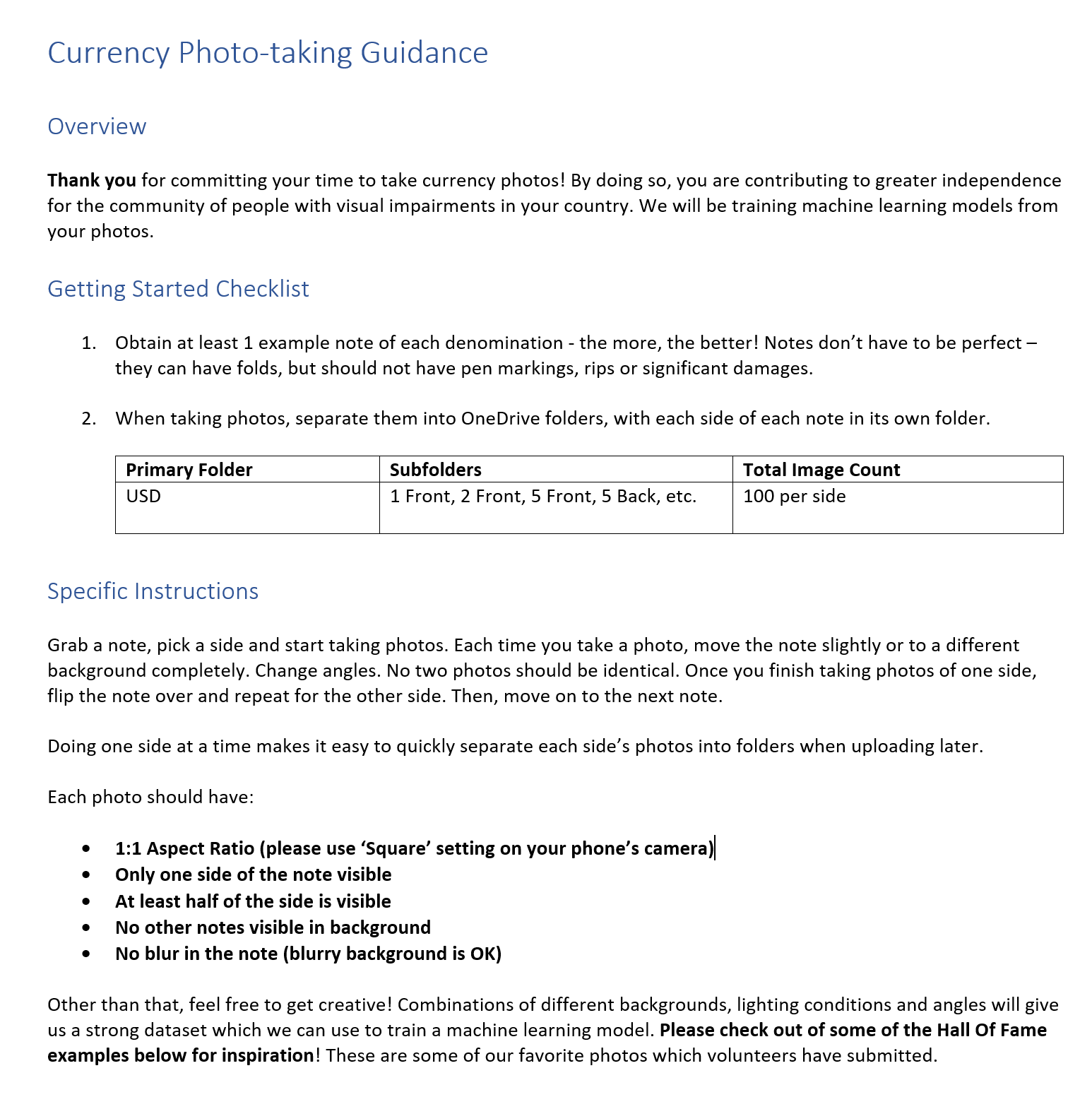}
\end{figure}

\begin{figure}[H]
\centering
\includegraphics[width=0.7\textwidth]{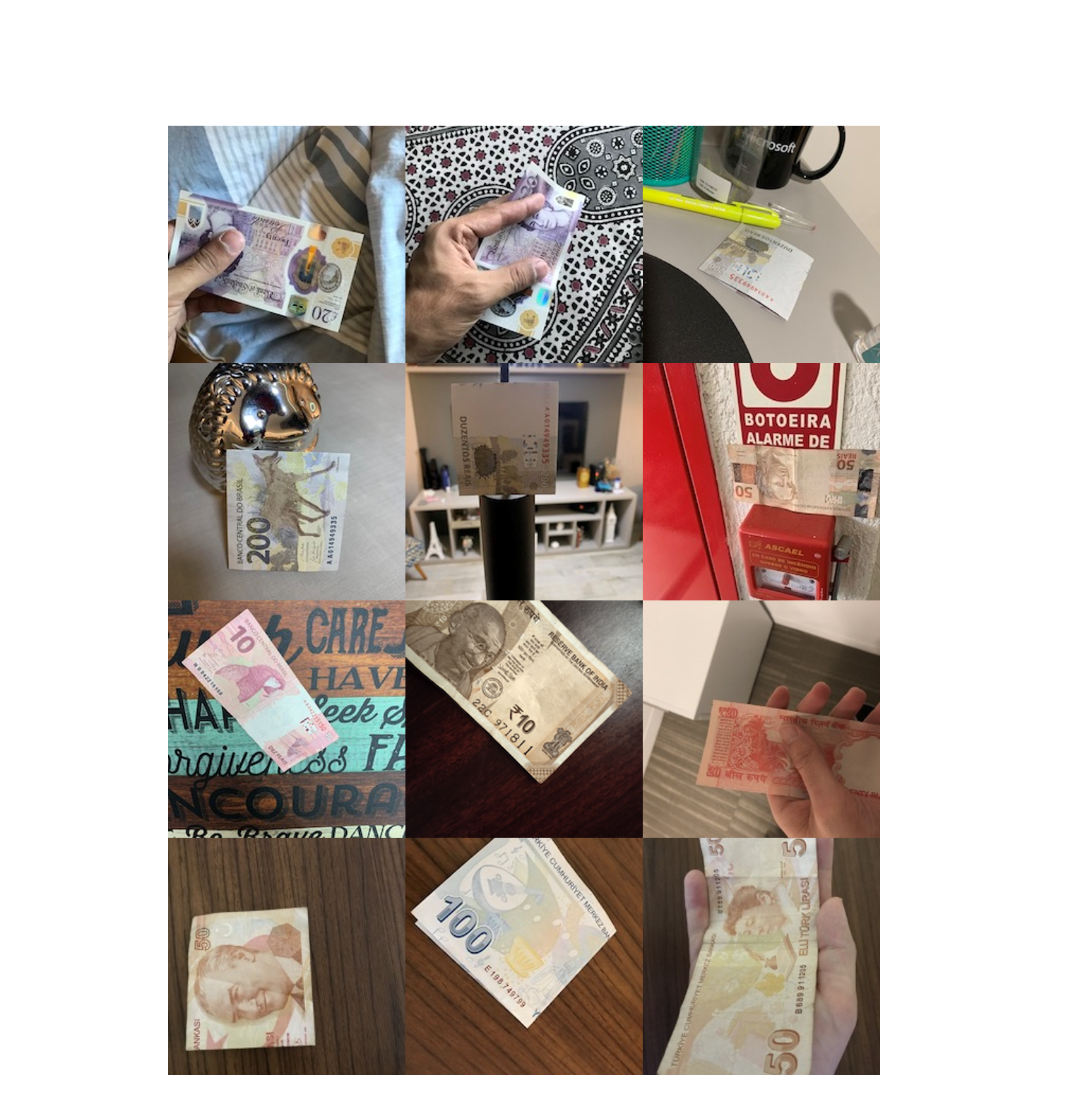}
\end{figure}

\begin{figure}[H]
\centering
\includegraphics[width=0.7\textwidth]{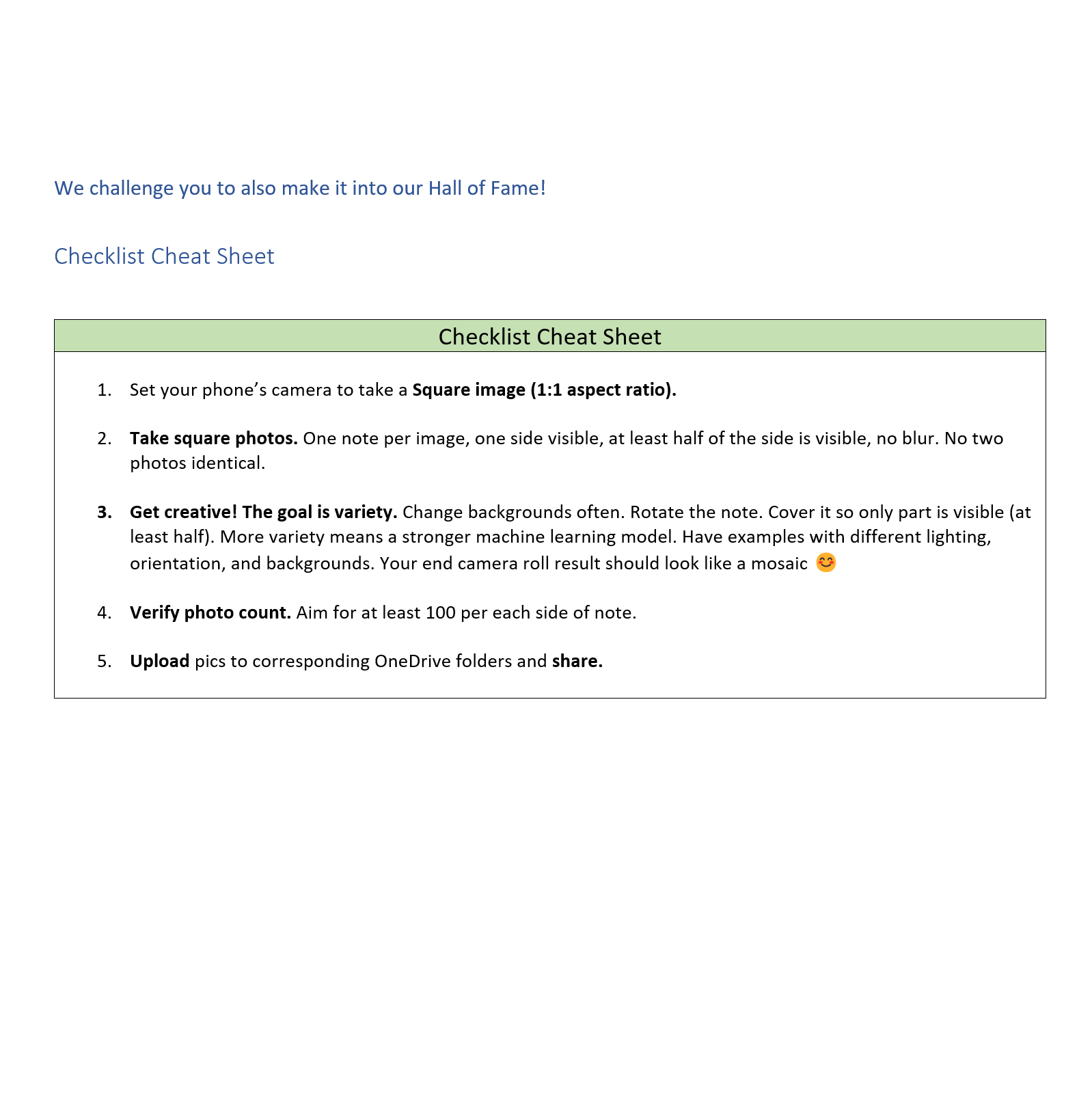}
\end{figure}

\section{Appendix B} \label{appendix-b}

List of classes in the BankNote-Net dataset, composed (in order) of a currency Forex abbreviation, denomination, face (front or back) and, if applicable, variation of banknote.

\begin{figure}[H]
\centering
\includegraphics[width=\textwidth]{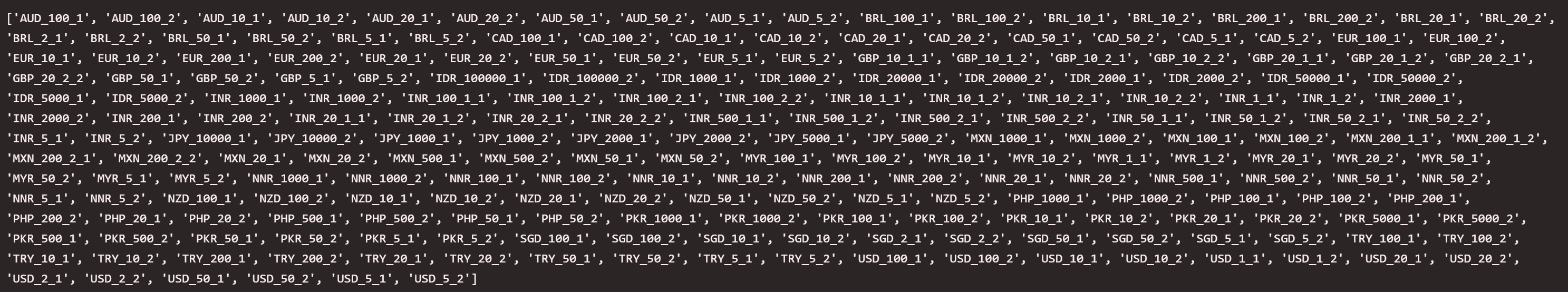}
\end{figure}

\end{document}